%% file: main.tex
\documentclass[12pt, titlepage]{report}
\usepackage[a4paper, top=1in, bottom=1in, right=1in, left=1.5in]{geometry}
\usepackage[utf8]{inputenc}
\usepackage{setspace}
\usepackage{hyphenat}
\usepackage{fancyhdr}
\usepackage{mathtools}
\usepackage{amsmath}
\usepackage[algoruled]{algorithm2e}
\everymath{\displaystyle}

\usepackage{graphicx}
\graphicspath{ {images/} }

\makeatletter
\def\hlinewd#1{%
  \noalign{\ifnum0=`}\fi\hrule \@height #1 \futurelet
   \reserved@a\@xhline}
\makeatother

\begin{document}
	\doublespacing
	
	\begin{titlepage}
		\begin{center}
			{\singlespacing \LARGE \uppercase{Deep Generative Networks for Sequence Prediction}\\[24pt]}
			
			{\Large Markus B. Beissinger\\[12pt]}
			
			\uppercase{a thesis}\\[12pt]					
			in\\			
			{\Large Computer and Information Science\\[36pt]}
			
			Presented to the Faculties of the University of Pennsylvania in Partial Fulfillment of the Requirements for the Degree of Master of Science in Engineering\\[12pt]
			
			2017\\[60pt]
			
			\singlespacing
			\makebox[2.5in]{\hrulefill}\\
			Lyle H. Ungar\\
			Supervisor of Thesis\\[48pt]
			
			\makebox[2.5in]{\hrulefill}\\
			Lyle H. Ungar\\
			Graduate Group Chairperson
			
			\doublespacing
		\end{center}
	\end{titlepage}
	
	\pagenumbering{roman}

	\chapter*{Acknowledgements}
	I would like to thank Lyle Ungar for his invaluable insights and methods of thinking throughout advising this thesis. I would also like to thank Mark Liberman (CIS, Linguistics, University of Pennsylvania) and Mitch Marcus (CIS, University of Pennsylvania) for their work on the thesis committee, and Li Yao and Sherjil Ozair (Computer Science and Operations Research, Universit\'{e} de Montr\'{e}al) for help with the GSN starter code.
	
	\chapter*{Abstract}
	\input{chapters/abstract}
	
	\tableofcontents
	\listoffigures
	\listoftables
	
	\chapter{Introduction}
	\pagenumbering{arabic}
	\input{chapters/introduction}
	
	\chapter{Related Work}
	\input{chapters/related_work}
	
	\chapter{Background: Deep Architectures}
	\input{chapters/background}

	\chapter{General Methodology: Deep Generative Stochastic Networks (GSN)}
	\input{chapters/general_methodology}

	\chapter{Model 1: Temporal GSN (TGSN)}
	\input{chapters/model1}

	\chapter{Model 2: Recurrent GSN (RNN-GSN)}	
	\input{chapters/model2}

	\chapter{Model 3: Sequence Encoder Network (SEN)}	
	\input{chapters/model3}

	\chapter{Discussion of Results}
	\input{chapters/discussion}

	\chapter{Conclusion}
	\input{chapters/conclusion}

	\bibliographystyle{plain}
	\bibliography{reference/bibliography}

\end{document}

%% file: chapters/abstract.tex
This thesis investigates unsupervised time series representation learning for sequence prediction problems, i.e. generating nice-looking input samples given a previous history, for high dimensional input sequences by decoupling the static input representation from the recurrent sequence representation. We introduce three models based on Generative Stochastic Networks (GSN) for unsupervised sequence learning and prediction. GSNs are a probabilistic generalization of denoising auto-encoders that learn unsupervised hierarchical representations of complex input data, while being trainable by backpropagation.

The first model, the Temporal GSN (TGSN), uses the latent state variables \(H\) learned by the GSN to reduce input complexity such that learning the representations \(H\) over time becomes linear. This means a simple linear regression step \(H \rightarrow H\) can encode the next set of latent state variables describing the input data in the sequence, learning \(P(H_{t+1}|H_{t-m}^t)\) for an arbitrary history, or context, window of size \(m\).

The second model, the Recurrent GSN (RNN-GSN), uses a Recurrent Neural Network (RNN) to learn the sequences of GSN parameters \(H\) over time. By having the progression of \(H\) learned by an RNN instead of through regression like the TGSN, this model can learn sequences with arbitrary time dependencies.

The third model, the Sequence Encoding Network (SEN), is a novel framework for learning deep sequence representations. It uses a hybrid approach of stacking alternating reconstruction generative network layers with recurrent layers, allowing the model to learn a deep representation of complex time dependencies.

Experimental results for these three models are presented on pixels of sequential handwritten digit (MNIST) data, videos of low-resolution bouncing balls, and motion capture data \footnote{code can be found at: https://github.com/mbeissinger/recurrent\_gsn}. The main contribution of this thesis is to provide evidence that GSNs are a viable framework to learn useful representations of complex sequential input data, and to suggest a new framework for deep generative models to learn complex sequences by decoupling static input representations from dynamic time dependency representations.

%% file: chapters/introduction.tex
Deep learning research has grown in popularity due to its ability to form useful feature representations of highly complex input data. Useful representations are those that disentangle the factors of variation of input data, preserving the information that is ultimately useful for the given machine learning task. In practice, these representations are used as input features to other algorithms, where in the past features would have been constructed by hand. Deep learning frameworks (especially deep convolutional neural networks \cite{lenet5}) have had recent successes for supervised learning of representations for many tasks, creating breakthroughs for both speech and object recognition \cite{seide11, krizhevsky12}.

Unsupervised learning of representations, however, has had slower progress. These models, mostly Restricted Boltzmann Machines (RBM) \cite{hinton06}, auto-encoders \cite{alain12}, and sparse-coding variants \cite{ranzato07}, suffer from the difficulty of marginalizing across an intractable number of configurations of random variables (observed, latent, or both). Each plausible configuration of latent and observed variables would be a mode in the distribution of interest \(P(X,H)\) or \(P(X)\) directly, and current methods of inference or sampling are forced to make strong assumptions about these distributions. Recent advances on the generative view of denoising auto-encoders and generative stochastic networks \cite{gsn} have alleviated this difficulty by simply learning a local Markov chain transition operator through backpropagation, which is often unimodal (instead of parameterizing the data distribution directly, which is multi-modal). This approach has opened up unsupervised learning of deep representations for many useful tasks, including sequence prediction. Unsupervised sequence prediction and labeling remains an important problem for artificial intelligence (AI), as many types of input data, such as language, video, etc., naturally form sequences and the vast majority is unlabeled.

This thesis will cover four main topics:
\begin{itemize}
	\item Chapter 3 provides an overview of deep architectures - a background on representation learning from probabilistic and direct encoding viewpoints. Recent work on generative viewpoints will be discussed as well, showing how denoising auto-encoders can solve the multi-modal problem via learning a Markov chain transition operator.
	\item Chapter 4 introduces Generative Stochastic Networks - recent work generalizing the denoising auto-encoder framework into GSNs will be explained, as well as how this can be extended to sequence prediction tasks.
	\item Chapters 5, 6, and 7 describe models using GSNs to learn complex sequential input data.
	\item Chapter 8 discusses the results of these three models and baselines and why they are able to use deep representations to learn sequence data.
\end{itemize}

%% file: chapters/related_work.tex
Due to the success of deep architectures on highly complex input data, applying deep architectures to sequence prediction tasks has been studied extensively in literature. RBM variants have been the most popular for applying deep learning models to sequential data.

\emph{Temporal RBM (TRBM)} \cite{sutskever06} is one of the first frameworks of non-linear sequence models that are more powerful than traditional Hidden Markov models or linear systems. It learns multilevel representations of sequential data by adding connections from previous states of the hidden and visible units to the current states. When the RBM is known, the TRBM learns the dynamic biases of the parameters from one set of states to the next. However, inference over variables is still exponentially difficult.

\emph{Recurrent Temporal RBMs (RTRBMs)} \cite{sutskever08} are an extension of the TRBM. They add a secondary learned latent variable \(H'\) that serves to reduce the number of posterior probabilities needed to consider during inference through a learned generative process. Exact inference can be done easily and gradient learning becomes almost tractable.

\emph{Temporal Convolution Machines (TCMs)} \cite{lockett09} also build from TRBMs. They make better use of prior states by allowing the time-varying bias of the underlying RBM to be a convolution of prior states with any function. Therefore, the states of the TCM can directly depend on arbitrarily distant past states. This means the complexity of the hidden states are reduced, as a complex Markov sequence in the hidden layer is not necessary. However, inference is still difficult.

\emph{RNN-RBM} \cite{lewandowski12} is similar to the RTRBM. The RNN-RBM adds a recursive neural network layer that acts as a dynamic state variable \(u\) which is dependent on the current input data and the past state variable. This state variable is what then determines the bias parameters of the next RBM in the sequence, rather than just a regression from the latents \(H\).

\emph{Sequential Deep Belief Networks (SDBNs)} \cite{andrew12, andrew13}  is a series of stacked RBMs that have a Markov interaction over time between each corresponding hidden layer. Rather than adjusting the bias parameters dynamically like TRBMs, this approach learns a Markov transition between the hidden latent variables over time. This allows the hidden layer to model any dependencies between time frames of the observations.

\emph{Recursive Neural Networks (RNNs)} \cite{socher11} are a slightly different framework used for sequence labeling in parsing natural language sentences or parsing natural scene images that have recursive structures. RNNs define a neural network that takes two possible input vectors (such as adjoining words in a sentence) and produces a hidden representation vector as well as a prediction score of the representation being the correct merging of the two inputs. These hidden representation vectors can be fed recursively into the RNN to calculate the highest probability recursive structure of the input sequence. RNNs are therefore a supervised algorithm.

Past work has also compared a deep architecture, Sentence-level Likelihood Neural Nets (SLNN), with traditional Conditional Random Fields (CRF) for sequence labeling tasks of Named Entity Recognition and Syntactic chunking \cite{wang13}. Wang et al. found that non-linear deep architectures, compared to linear CRFs, are more effective in low dimensional continuous input spaces, but not in high-dimensional discrete input spaces. They also confirm that distributional representations can be used to achieve better generalization.

While many of these related works perform well on sequential data such as video and language, all of them (except for the RTRBM) still struggle with inference due to the nature of RBMs. Using these sequential techniques on GSNs, which are easy to sample from and perform inference, have not yet been studied.

%% file: chapters/background.tex
Traditional machine learning algorithms' performance depend heavily on the particular features of the data chosen as inputs. For example, document classification (such as marking emails as spam or not spam) can be performed by breaking down the input document into bag-of-words or n-grams as features. Choosing the correct feature representation of input data, or feature engineering, is a way to bring prior knowledge of a domain to increase an algorithm's computational performance and accuracy. To move towards general artificial intelligence, algorithms need to be less dependent on this feature engineering and better learn to identify the explanatory factors of input data on their own \cite{bengio12}.

\section{Representation Learning}
Deep learning frameworks (also known as deep architectures) move in this direction by capturing a good representation of input data by using compositions of non-linear transformations. A good representation can be defined as one that disentangles underlying factors of variation for input data \cite{bengio13}. Deep learning frameworks can find useful abstract representations of data across many domains: it has had great commercial success powering most of Google and Microsoft's current speech recognition, image classification, natural language processing, object recognition, etc. Facebook is also planning on using deep learning to understand its users\footnote{http://www.technologyreview.com/news/519411/facebook-launches-advanced-ai-effort-to-find-meaning-in-your-posts/}. Deep learning has been so impactful in industry that MIT Technology Review named it as a top-10 breakthrough technology of 2013\footnote{http://www.technologyreview.com/featuredstory/513696/deep-learning/}.

The central idea to building a deep architecture is to learn a hierarchy of features one level at a time where the input to one computational level is the output of the previous level for an arbitrary number of levels. Otherwise, shallow representations (such as regression or support vector machines) go directly from input data to output classification.

One loose analogue for deep architectures is neurons in the brain (a motivation for artificial neural networks) - the output of a group of neurons is agglomerated as the input to more neurons to form a hierarchical layer structure. Each layer \(N\) is composed of \(h\) computational nodes that connect to each computational node in layer \(N+1\).

\begin{figure}[h!]
  \centering
    \includegraphics[width=0.6\textwidth]{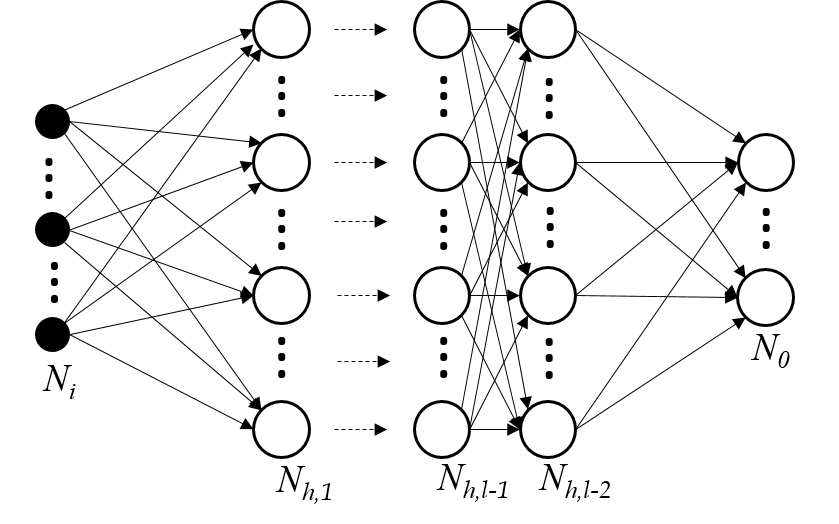}
\caption{An example deep architecture.}
\end{figure}

\section{Interpretations of Deep Architectures}
There are two main ways to interpret the computation performed by these layered deep architectures:

\begin{itemize}
\item \emph{Probabilistic graphical models} have nodes in each layer that are considered as latent random variables. In this case, the probability distribution of the input data \(x\) and the hidden latent random variables \(h\) that describe the input data in the joint distribution \(p(x,h)\) are the model. These latent random variables describe a distribution over the observed data.
\item \emph{Direct encoding models} have nodes in each layer that are considered as computational units. This means each node \(h\) performs some computation (normally nonlinear, such as a sigmoidal function, hyperbolic tangent, or rectifier linear unit) given its inputs from the previous layer.
\end{itemize}

To illustrate, principal component analysis (PCA) is a simple feature extraction algorithm that can span both of these interpretations. PCA learns a linear transform \(h = f(x) = W^T x + b\) where \(W\) is a weight matrix for the inputs \(x\) and \(b\) is a bias term. The columns of the \(dx \times dh\) matrix \(W\) form an orthogonal basis for the \(dh\) orthogonal directions of greatest variance in the input training data \(x\). The result is \(dh\) decorrelated features that make representation layer \(h\). 

\begin{figure}[h!]
  \centering
    \includegraphics[width=0.65\textwidth]{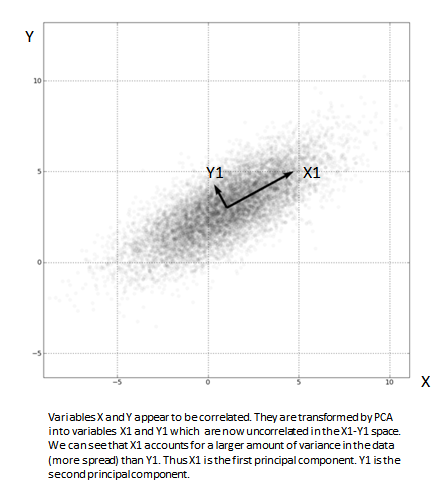}
\caption[PCA]{Principal component analysis\footnotemark{}.}
\end{figure}
\footnotetext{http://www.simafore.com/Portals/64283/images/principal-component-analysis-simple-explanation.png}

From a probabilistic viewpoint, PCA is simply finding the principal eigenvectors of the covariance matrix of the data. PCA finds which features of the input data can explain away the most variance in the data\cite{bach05}. From an encoding viewpoint, PCA is performing a linear computation over the input data to form a hidden representation h that has a lower dimensionality than the data.

Note that because PCA is a linear transformation of the input \(x\), it cannot really be stacked in layers because the composition of linear operations is just another linear operation. There would be no abstraction benefit of multiple layers. To show these two methods of analysis, this section will examine stacking Restricted Boltzmann Machines (RBM) from a probability viewpoint and nonlinear auto-encoders from a direct encoding viewpoint.

\subsection{Probabilistic models: Restricted Boltzmann Machine (RBM)}
A Boltzmann machine is a network of symmetrically-coupled binary random variables or units. This means that it is a fully-connected, undirected graph. This graph can be divided into two parts:

\begin{enumerate}
\item The visible binary units \(x\) that make up the input data and
\item The hidden or latent binary units \(h\) that explain away the dependencies between the visible units \(x\) through their mutual interactions.
\end{enumerate}

Boltzmann machines describe this pattern of interaction through the distribution over the joint space \([x,h]\) with the energy function: 
\[\varepsilon_\Theta^{BM} (x,h) = -\frac{1}{2} x^T Ux - \frac{1}{2} h^T Vh - x^T Wh - b^T x - d^T h\]
Where the model parameters \(\Theta\) are \(\left\{U,V,W,b,d\right\}\).

Evaluating conditional probabilities over this fully connected graph ends up being an intractable problem. For example, computing the conditional probability of a hidden variable given the visibles, \(P(h_i | x)\), requires marginalizing over all the other hidden variables. This would be evaluating a sum with \(2dh - 1\) terms.

However, we can restrict the graph from being fully connected to only containing the interactions between the visible units \(x\) and hidden units \(h\). 

\begin{figure}[h!]
  \centering
    \includegraphics[width=0.4\textwidth]{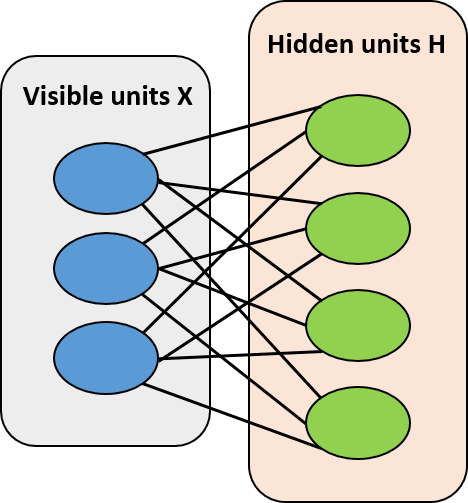}
\caption{A Restricted Boltzmann Machine.}
\end{figure}

This gives us an RBM, which is a \emph{bipartite} graph with the visible and hidden units forming distinct layers. Calculating the conditional distribution \(P(h_i | x)\) is readily tractable and factorizes to: 
\[P(h | x) = \prod_i P(h_i | x)\]
\[P(h_i = 1 | x) = sigmoid \left( \sum_j W_{ji} x_j + d_i \right)\]

Very successful deep learning algorithms stack multiple RBMs together, where the hidden units \(h\) from the visible input data \(x\) become the new input data for another RBM for an arbitrary number of layers. 

\begin{figure}[h!]
  \centering
    \includegraphics[width=0.2\textwidth]{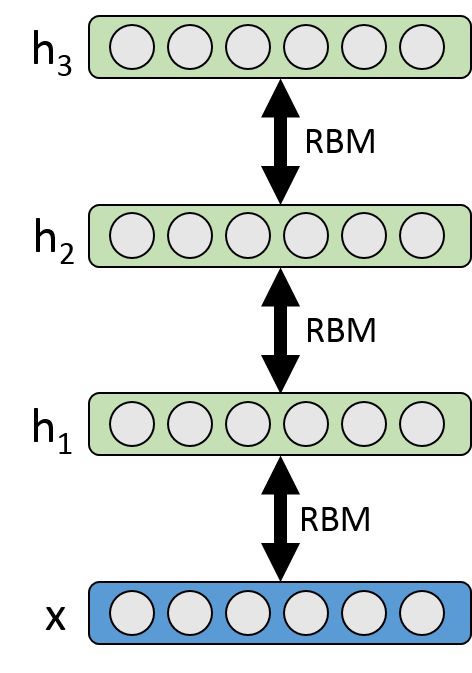}
\caption{Stacked RBM.}
\end{figure}

There are a few drawbacks to the probabilistic approach to deep architectures:
\begin{enumerate}
\item The posterior distribution \(P(h_i | x)\) becomes incredibly complicated if the model has more than a few interconnected layers. We are forced to resort to sampling or approximate inference techniques to solve the distribution, which has computational and approximation error prices.
\item Calculating this distribution over latent variables still does not give a usable feature vector to train a final classifier to make this algorithm useful for AI tasks. For example, the calculations of these hidden distributions explain the variations over the handwriting digit recognition problem, but they do not give a final classification of a number. Actual feature values are normally derived from the distribution, taking the latent variable's expected value, which are then used as the input to a normal machine learning classifier, such as logistic regression.
\end{enumerate}

\subsection{Direct encoding models: auto-encoder}
To get around the problem of deriving useful feature values, an auto-encoder is a non-probabilistic alternative approach to deep learning where the hidden units produce usable numeric feature values. An auto-encoder directly maps an input \(x\) to a hidden layer \(h\) through a parameterized closed-form equation called an encoder. Typically, this encoder function is a nonlinear transformation of the input to \(h\) in the form:
\[f_\Theta (x) = s_f (Wx + b)\]

This resulting transformation is the feature-vector or representation computed from input \(x\).
Conversely, a decoder function is then used to map from this feature space \(h\) back to the input space, which results in a reconstruction \(x'\). This decoder is also a parameterized closed-form equation that is a nonlinear function undoing the encoding function:
\[g_\Theta (h) = s_g (W' h + d)\]

In both cases, the nonlinear function s is normally an element-wise sigmoid, hyperbolic tangent nonlinearity, or rectifier linear unit.

Thus, the goal of an auto-encoder is to minimize a loss function over the reconstruction error given the training data. Model parameters \(\Theta\) are \(\left\{W,b,W',d\right\}\), with the weight matrix \(W\) most often having tied weights such that \(W' = W^T\) .

Stacking auto-encoders in layers is the same process as with RBMs.
\begin{figure}[h!]
  \centering
    \includegraphics[width=0.8\textwidth]{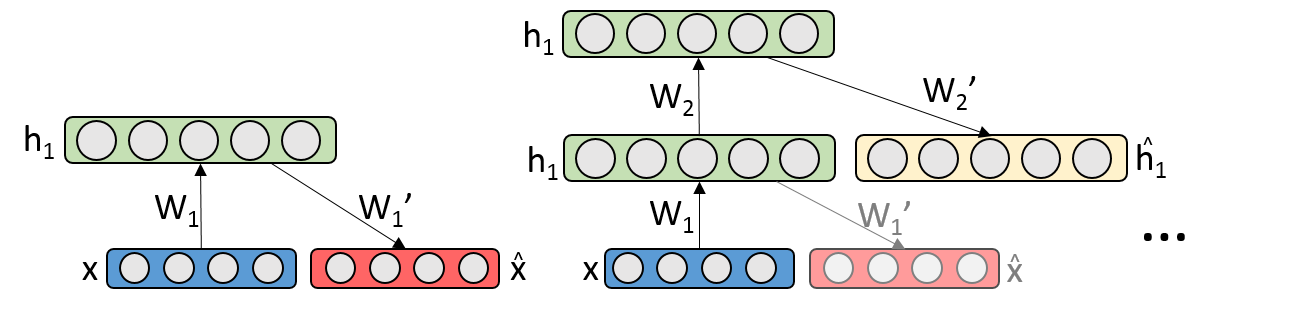}
\caption{Stacked auto-encoder.}
\end{figure}

One disadvantage of auto-encoders is that they can easily memorize the training data (i.e., find the model parameters that map every input seen to a perfect reconstruction with zero error) given enough hidden units \(h\). To combat this problem, regularization is necessary, which gives rise to variants such as sparse auto-encoders, contractive auto-encoders, or denoising auto-encoders.

A practical advantage of auto-encoder variants is that they define a simple, tractable optimization objective that can be used to monitor progress.

\section{Denoising Auto-encoders}
Denoising auto-encoders \cite{bengio13a, vincent08, alain12} are a class of direct encoding models that use synthetic noise over the inputs through a corruption process during training to prevent overfitting and simply learning the identity function. Given a known corruption process \(C(\widetilde{X}|X)\) to corrupt an observed variable \(X\), the denoising auto-encoder learns the reverse conditional \(P(X|\widetilde{X})\). Combining this estimator with the known corruption process \(C\), it can recover a consistent estimator of \(P(X)\) through a Markov chain that alternates sampling from \(C(\widetilde{X}|X)\) and \(P(X|\widetilde{X})\). The basic algorithm is as follows:

\begin{algorithm}[h!]
	\KwIn{training set \(D\) of examples \(X\), a corruption process \(C(\widetilde{X}|X)\), and a conditional distribution \(P_\Theta(X|\widetilde{X})\) to train.}
	\While{training not converged}{
		sample training example \(X \sim D\)\;
		sample corrupted input \(\widetilde{X} \sim C(\widetilde{X}|X)\)\;
		use (\(X\),\(\widetilde{X}\)) as an additional training example towards minimizing the expected value of \(-\log P_\Theta (X|\widetilde{X})\), e.g., by a gradient step with respect to \(\Theta\) in the encoding/decoding function\;
	}
	\caption{ Generalized Denoising Auto-encoder Training Algorithm }
\end{algorithm}

The reconstruction distribution \(P(X|\widetilde{X})\) is easier to learn than the true data distribution \(P(X)\) because \(P(X|\widetilde{X})\) is often dominated by a single or few major modes, where the data distribution \(P(X)\) would be highly multimodal and complex. Recent works \cite{alain12, bengio13a} provide proofs that denoising auto-encoders with arbitrary variables (discrete, continuous, or both), an arbitrary corruption (Gaussian or other; not necessarily asymptotically small), and an arbitrary loss function (as long as it is viewed as a log-likelihood) estimate the score (derivative of the log-density with respect to the input) of the observed random variables.

Another key idea presented in Bengio et al. \cite{bengio13a} is walkback training. The walkback process generates additional training examples through a pseudo-Gibbs sampling process from the current denoising auto-encoder Markov chain for a certain number of steps. These additional generated \((X,\widetilde{X})\) pairs from the model decrease training time by actively correcting spurious modes (regions of the input data that have been insufficiently visited during training, which may therefore be incorrect in the learned reconstruction distribution). Both increasing the number of training iterations and increasing corruption noise alleviate spurious modes, but walkbacks are the most effective.
\begin{algorithm}[h!]
	\KwIn{A given training example \(X\), a corruption process \(C(\widetilde{X}|X)\), and the current model's reconstruction conditional distribution \(P_\Theta(X|\widetilde{X})\). It also has a hyper-parameter \(p\) that controls the number of generated samples.}
	\KwOut{A list \(L\) of additional training examples \(\widetilde{X}^*\).}
	\(X^* \leftarrow X\), \(L \leftarrow []\)\;
	Sample \(\widetilde{X}^* \sim C(\widetilde{X}|X^*)\)\;
	Sample \(u \sim\) Uniform\((0,1)\)\;
	\While{\(u < p\)}{
		Append \(\widetilde{X}^*\) to \(L\), so \((X,\widetilde{X}^*)\) will be an additional training example for the denoising auto-encoder.\;
		Sample \(X^* \sim P_\Theta(X|\widetilde{X}^*)\)\;
		Sample \(\widetilde{X}^* \sim C(\widetilde{X}|X^*)\)\;
		Sample \(u \sim\) Uniform\((0,1)\)\;
	}
	Append \(\widetilde{X}^*\) to \(L\)\;
	Return \(L\)\;
	\caption{ Walkback Training Algorithm for Denoising Auto-encoders }
\end{algorithm}

%% file: chapters/general_methodology.tex
Generative stochastic networks are a generalization of the denoising auto-encoder and help solve the problem of mixing between many modes as outlined in the Introduction. Each model presented in this thesis uses the GSN framework for learning a more useful abstraction of the input distribution \(P(X)\).

\section{Generalizing denoising auto-encoders}

Denoising auto-encoders use a Markov chain to learn a reconstruction distribution \(P(X|\widetilde{X})\) given a corruption process \(C(\widetilde{X}|X)\) for some data \(X\). Denoising auto-encoders have been shown as generative models \cite{bengio13a}, where the Markov chain can be iteratively sampled from:
\begin{align*}
 &X_t \sim P_\Theta(X|\widetilde{X}_{t-1})\\
 &\widetilde{X}_t \sim C(\widetilde{X}|X_t)
\end{align*}

As long as the learned distribution \(P_{\Theta_n}(X|\widetilde{X})\) is a consistent estimator of the true conditional distribution \(P(X|\widetilde{X})\) and the Markov chain is ergodic, then as \(n \rightarrow \infty\), the asymptotic distribution \(\pi_n(X)\) of the generated samples from the denoising auto-encoder converges to the data-generating distribution \(P(X)\) (proof provided in Bengio et al. \cite{bengio13a}).

\subsection{Easing restrictive conditions on the denoising auto-encoder}

A few restrictive conditions are necessary to guarantee ergodicity of the Markov chain - requiring \(C(\widetilde{X}|X) > 0\) everywhere that \(P(X) > 0\). Particularly, a large region \(V\) containing any possible \(X\) is defined such that the probability of moving between any two points in a single jump \(C(\widetilde{X}|X)\) must be greater than 0. This restriction requires that \(P_{\Theta_n}(X|\widetilde{X})\) has the ability to model every mode of \(P(X)\), which is a problem this model was meant to avoid.

To ease this restriction, Bengio et al. \cite{gsn} proves that using a \(C(\widetilde{X}|X)\) that only makes small jumps allows \(P_{\Theta}(X|\widetilde{X})\) to model a small part of the space \(V\) around each \(\widetilde{X}\). This weaker condition means that modeling the reconstruction distribution \(P(X|\widetilde{X})\) would be easier since it would probably have fewer modes. 

However, the jump size \(\sigma\) between points must still be large enough to guarantee that one can jump often enough between the major modes of \(P(X)\) to overcome the deserts of low probability: \(\sigma\) must be larger than half the largest distance of low probability between two nearby modes, such that \(V\) has at least a single connected component between modes. This presents a tradeoff between the difficulty of learning \(P_{\Theta}(X|\widetilde{X})\) and the ease of mixing between modes separated by this low probability desert.

\subsection{Generalizing to GSN}

While denoising auto-encoders can rely on \(X_t\) alone for the state of the Markov chain, GSNs introduce a latent variable \(H_t\) that acts as an additional state variable in the Markov chain along with the visible \(X_t\) \cite{gsn}:
\begin{align*}
 &H_{t+1} \sim P_{\Theta_1}(H|H_t, X_t)\\
 &X_{t+1} \sim  P_{\Theta_2}(X|H_{t+1})
\end{align*}
The resulting computational graph takes the form:

\begin{figure}[h!]
  \centering
    \includegraphics[width=0.6\textwidth]{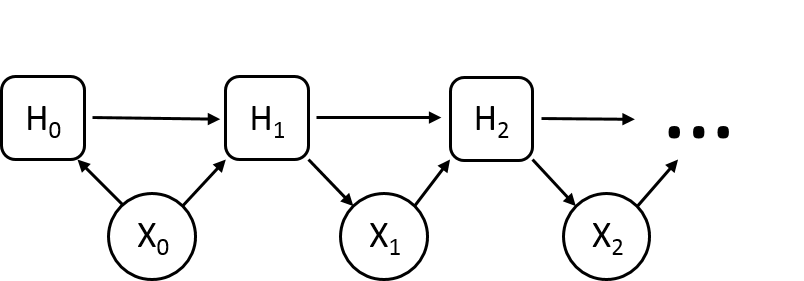}
\caption{GSN computational graph.}
\end{figure}
The latent state variable \(H\) can be equivalently defined as \(H_{t+1} = f_{\Theta_1}(X_t,Z_t,H_t)\), a learned function \(f\) with an independent noise source \(Z_t\) such that \(X_t\) cannot be reconstructed exactly from \(H_{t+1}\). If \(X_t\) could be recovered from \(H_{t+1}\), the reconstruction distribution would simply converge to the Dirac at \(X\). Denoising auto-encoders are therefore a special case of GSNs, where \(f\) is fixed instead of learned.

GSNs also use the notion of walkback to aid training. The resulting Markov chain of a GSN is inspired by Gibbs sampling, but with stochastic units at each layer that can be backpropagated \cite{rezende14}.

\begin{figure}[h!]
  \centering
    \includegraphics[width=0.8\textwidth]{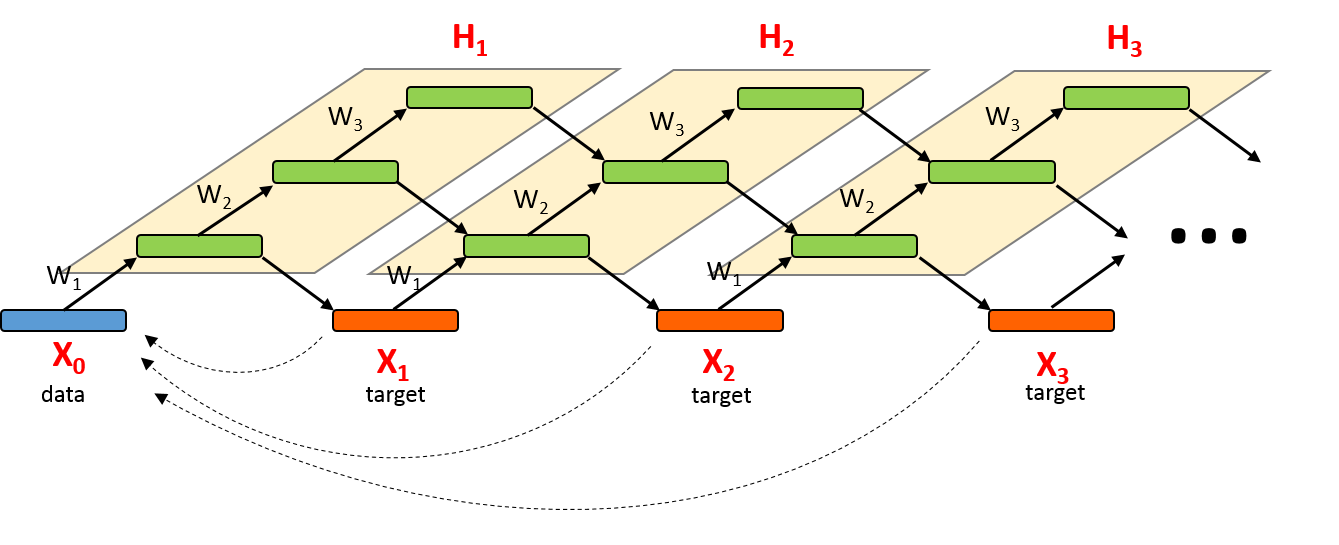}
\caption{Unrolled GSN Markov chain.}
\end{figure}

\section{Extension to recurrent deep GSN}

Similar to RBMs, sequences of GSNs can be learned by a recurrent step in the parameters or latent states over the sequence of input variables. This approach works because deep architectures help solve the multi-modal problem of complex input data explained in the Introduction and can easily mix between many modes.

The main mixing problem comes from the complicated data manifold surfaces of the input space; transitioning from one MNIST digit to the next in the input space generally looks like a messy blend of the two numbers in the intermediate steps. As more layers are learned, more abstract features lead to better disentangling of the input data, which ends up unfolding the manifolds to fill a larger part of the representation space. Because these manifolds become closer together, Markov Chain Monte Carlo (MCMC) sampling between them moves more easily between the modes of the input data and creates a much better mixing between the modes.

\begin{figure}[h!]
  \centering
    \includegraphics[width=1.0\textwidth]{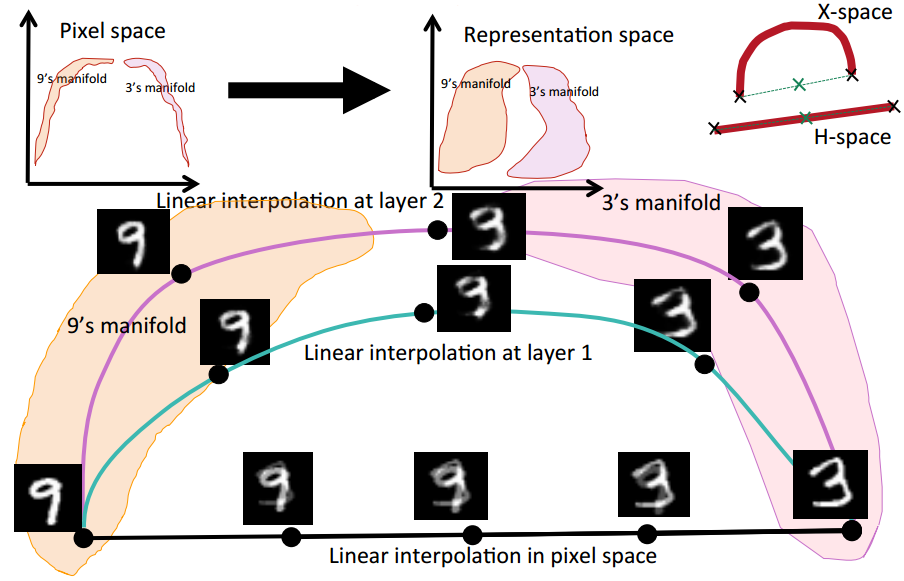}
\caption{Better mixing via deep architectures \cite{bengio_workshop}.}
\end{figure}

Because the data manifold space becomes less complicated at higher levels of abstraction, transitioning between them over time becomes much easier. This principle enables the models in the following three chapters to learn sequences of complex input data over time.

%% file: chapters/model1.tex
This first approach is most similar to Sequential Deep Belief Networks in that it learns a transition operator between hidden latent states \(H\). This model uses the power of GSN's to learn hidden representations that reduce the complexity of the input data space, making transitions between data manifolds at higher layers of representation much easier to model. Therefore, the transition step of learning \(H \rightarrow H\) over time should be less complicated (i.e. only needing a single linear regression step between hidden states). This model trains by alternating over two versions of the dataset:
\begin{enumerate}
\item A generative Gibbs sampling pass for \(k\) samples on each input in arbitrary order (for the GSN to learn the data manifolds)
\item A real time-sequenced order of the input to learn the regression \(H \rightarrow H\)
\end{enumerate}
Alternating between training the GSN parameters on the generative input sequence through Gibbs sampling and learning the hidden state transition operator on the real sequence of inputs allows the model to tune parameters quickly in an expectation-maximization style of training.

\section{Recurrent nature of deep GSNs}

While GSNs are inherently recurrent and depend on the previous latent and visible states to determine the current hidden state, \(H_{t} \sim P_{\Theta_1}(H|H_{t-1},X_{t-1})\), this ordering \(t\) is generated through the GSN Gibbs sampling process and does not reflect the real sequence of inputs over time. Using this sampling process, GSNs actively mix between modes that are close together in the input space, not the sequential space. For example, a GSN trained on MNIST data will learn to mix well between the modes of digits that look similar in the pixel space - sampling from the digit ``4" transitions to a ``9", etc.

\begin{figure}[h!]
  \centering
    \includegraphics[width=0.8\textwidth]{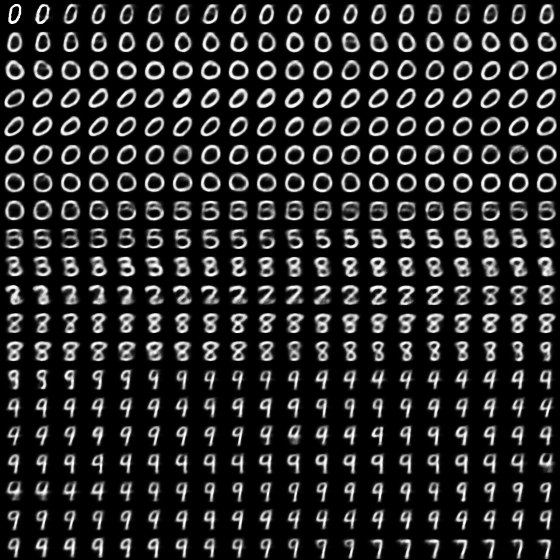}
\caption{Samples from GSN after 290 training epochs. Good mixing between major modes in the input space.}
\end{figure}

To learn transitions between sequential modes in the input space, both the sample step \(t\) from the GSN sampling process and the sequential input \(t_{real}\) from the input data sequence need to be utilized.

\section{Algorithm}

\begin{figure}[h!]
  \centering
    \includegraphics[width=0.8\textwidth]{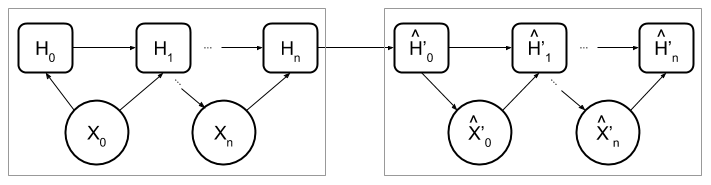}
\caption{Temporal GSN architecture.}
\end{figure}

This model's training algorithm is similar to expectation maximization (EM) - first optimizing GSN parameters over the input data, then learning the transition \(H \rightarrow H\) parameters, and repeating. After the initial GSN pass over the data, training the GSN parameters becomes more powerful as the reconstruction cost of the current input as well as the next predicted input are both used for computing the gradients.

One difficulty with this training setup is that the first few epochs will have the reconstruction cost after predicting the transition operator be incorrect until the GSN parameters are warmed up. The GSNs converge slowly and can get stuck in bad configurations due to the regression step (which is simple linear regression) being trained on a not-yet-useful hidden representation of the complex input data. If the regression step is trained poorly, it affects the remaining GSN parameter training steps by providing bad sequential predictions for the GSN to attempt to reconstruct. In practice we recommend waiting until the GSN reconstruction cost starts to converge before applying the reconstruction cost of the predicted next step to the GSN training operation.

\begin{algorithm}[h!]
	\KwIn{training set \(D\) of examples \(X\) in sequential order, \(N\) layers, \(k\) walkbacks}
	Initialize GSN parameters \(\Theta_{GSN} = \) \{List(weights from one layer to the next), List(bias for layer))\}\;
	Initialize transition parameters \(\Theta_{transition}=\) \{List(weights from previous layer to current), List(bias for layer)\} \;
	\While{training not converged}{
		\For{each input \(X\)}{
			Sample GSN for \(k\) walkbacks, creating \(k^*(X,X_{recon})\) training pairs\;
			Transition from ending hidden states \(H\) to next predicted hidden states \(H'\) with transition parameters \(\Theta_{transition}\)\;
			Sample GSN again for \(k\) walkbacks, creating  \(k^*(X',X'_{recon})\) training pairs\;
			Train GSN parameters \(\Theta_{GSN}\) using these pairs, keeping \(\Theta_{transition}\) fixed\;
		}
		\For{each input \(X\)}{
			Sample GSN for \(k\) walkbacks, creating ending hidden states \(H\)\;
			Transition from ending hidden states \(H\) to next predicted hidden states \(H'\) with transition parameters \(\Theta_{transition}\)\;
			Sample GSN again for \(k\) walkbacks, creating the ending \((X',X'_{recon})\) pair\;
			Train transition parameters \(\Theta_{transition}\) with this pair, keeping \(\Theta_{GSN}\) fixed\;
		}
	}
	\caption{ Model 1 EM Algorithm }
\end{algorithm}

\section{Experimental setup}

This algorithm was tested on artificially sequenced MNIST handwritten digit data. The dataset was sequenced by ordering the inputs 0-9 repeating. The GSN uses hyperbolic tangent (tanh) activation with 3 hidden layers of 1500 nodes and sigmoidal activation for the visible layer. For the GSN, Gaussian noise is added pre- and post-activation with a mean of 0 and a sigma of 2, and input corruption noise is salt-and-pepper with p=0.4. Training was performed for 300 iterations over the input data using a batch size of 100, with a learning rate of 0.25, annealing rate of 0.995, and momentum of 0.5.

An interesting result is that the predicted reconstruction of the next digits appears to be close to the average of that digit, which can be explained because the training set of sequences was shuffled and re-ordered after every epoch from the pool of available digits.

\begin{figure}[h!]
  \centering
    \includegraphics[width=1.0\textwidth]{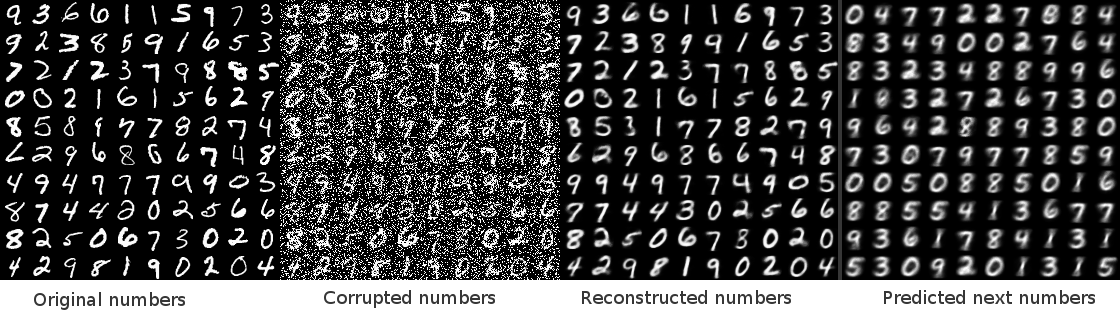}
\caption{Model 1 reconstruction of digits and predicted next digits after 300 iterations.}
\end{figure}

\begin{figure}[h!]
  \centering
    \includegraphics[width=0.5\textwidth]{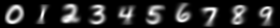}
\caption{Average MNIST training data by digit.}
\end{figure}

Differences between the predicted next number and the average number seem to occur when the GSN incorrectly reconstructs the original corrupted input. These results provide evidence that the original assumption is correct: the GSN learns representations that disentangle complex input data, which allows a simple regression step to predict the next input in a linear manner. A comparison of results is included in the Discussion.

Sampling in the input space is similar to a GSN:
\begin{figure}[h!]
  \centering
    \includegraphics[width=.8\textwidth]{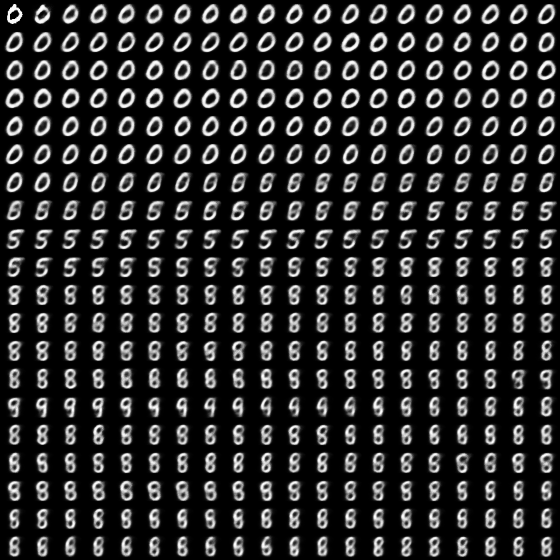}
\caption{Model 1 sampling after 90 training iterations; smooth mixing between major modes.}
\end{figure}

%% file: chapters/model2.tex
While the Temporal GSN (TGSN) works well predicting the next input digit given the current one, it is limited by its regression transition operators learning a linear mapping \(H \rightarrow H\) of a given context window. This inherently limits the length and complexity of sequences learnable in the latent space. The Recurrent GSN in this chapter introduces an additional recurrent latent parameter \(V\) to learn the sequence of GSN \(H\)s over time to tackle this problem.

\section{Aside: GSN as a recurrent network}
As Section 5.1 shows, GSNs inherently have a recurrent structure when the Gibbs chain is unrolled. Instead of using Gibbs sampling to generate inputs of the same class, a GSN can use the real time sequence of the input data to train its parameters with respect to the predicted reconstruction of the next input in the sequence. The GSN becomes a generative sequence prediction model rather than a generative data distribution model. This approach is not without drawbacks - the GSN loses its ability to utilize the walkback training principle for creating robust representations by actively seeking out spurious modes in the model. However, this drawback is mitigated with more input data.  Further, the GSN loses the ability to mix between modes of the input space. Instead, it mixes between modes of the sequence space - learning to transition to the next most likely input given the current and previous data.

Currently, GSNs use tied weights between layers to make backpropagation easier. However, this approach prohibits the hidden representation from being able to encode sequences. We must untie these weights to consider a GSN as an RNN variant, which makes training more difficult.

\begin{algorithm}[h!]
	\KwIn{training data \(X\) in sequential order, \(N\) layers, \(k \geq 2*N\) predictions}
	Initialize GSN parameters \(\Theta_{GSN} = \) \{List(weights from one layer to the next higher), List(weights from one layer to the next lower), List(bias for layer))\}\;
	\For{input data \(x\) received}{
		 Sample from GSN predicted \(x'\) to create a list of the next \(k\) predicted inputs\;
		Store these predictions in a memory buffer array of lists\;
		Use the current input \(x\) to train GSN parameters with respect to the list of predicted \(x'\) through backpropagation\;
	}
	\caption{ Untied GSN as an RNN }
\end{algorithm}

\subsection{Untied GSN on sequenced MNIST}

Using the same general training parameters with regards to noise, learning rate, annealing, momentum, epochs, hidden layers, and activation as the TGSN, untying the GSN parameters performs similarly with regards to binary cross-entropy as the TGSN on the artificially sequenced MNIST dataset. For the next immediate predicted number, it achieved a binary cross-entropy of 0.2318.  For the predicted number six iterations ahead, it achieved a binary cross-entropy of 0.2268. This cross-entropy is lower because six iterations ahead can utilize higher layers of representation in the GSN due to the way the computational graph is formed.

Even though cross-entropy is similar to the TGSN, reconstruction images paint a different picture. Due to untied weights taking longer to train, the next predicted digits appear worse than the averages produced from the TGSN over the same number of iterations. However, as the number of predictions ahead increases, the digits begin to look more like the averages. This could be explained by further predictions ahead utilizing the higher layers of representation based on the way the computational graph is formed.

\begin{figure}[h!]
  \centering
    \includegraphics[width=1.0\textwidth]{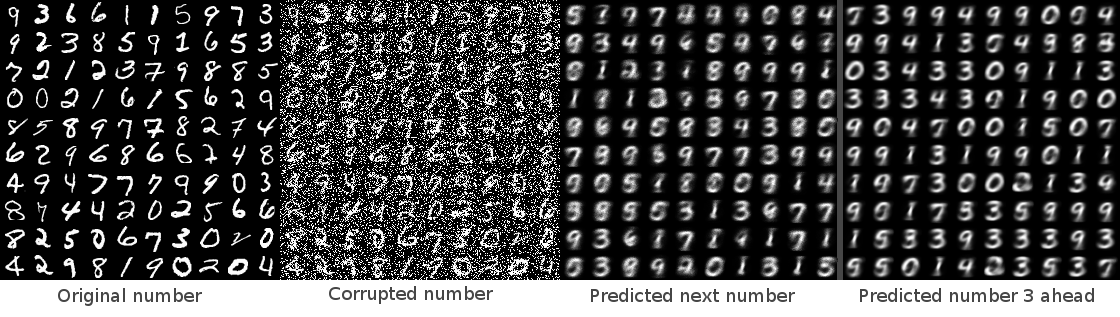}
\caption{Untied GSN reconstruction of predicted next digits and predicted digits 3 iterations ahead after 300 iterations.}
\end{figure}

Learning with untied weights is much slower, but still provides evidence that the hidden layers themselves can learn useful representations for complex input sequences. Looking at the generated samples in Figure 6.2 after 300 training iterations, mixing between sequential modes is evident as the samples appear to be generated in the same 0-9 order as the sequenced data.

\begin{figure}[h!]
  \centering
    \includegraphics[width=.8\textwidth]{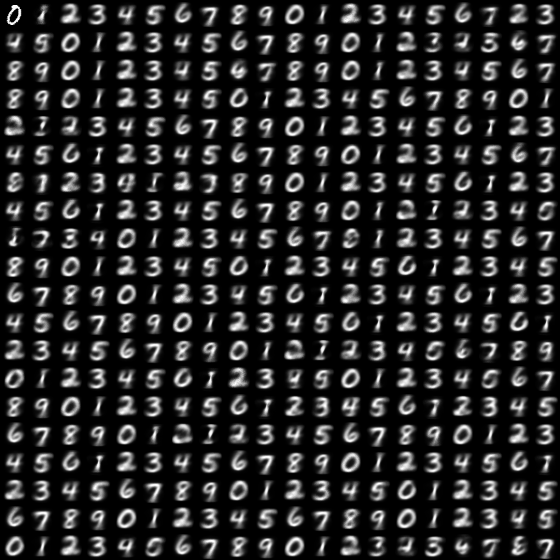}
\caption{Untied GSN sampling after 300 iterations.}
\end{figure}

The quality of images shown here encourage the use of separate parameters to decouple sequential learning from input representation learning.

\section{Extending the walkback procedure to sequenced inputs}

This online model loses the ability to use walkback training to reduce spurious modes. However, walkback could be generalized to the sequential case by sampling from possible variations of past hidden representations that could lead to the current input. Intuitively, this idea comes from the method of explaining current inputs with imperfect memory recall of past inputs. By sampling from the representation layer repeatedly, a series of potentially viable past representations that lead to the current input are created and used to train GSN parameters leading to the current input. This method uses past inputs as context to create viable variations of sequences in the representation space, which in turn acts to create more robust mixing between the modes in the sequence space.

The general process for creating sequential walkbacks described here is as follows:
\begin{algorithm}[h!]
	\For{k walkbacks}{
		Given input \(x\), take a backward step with the GSN using transposed weights and negated bias to create the previous hidden representation \(H\)\;
		Sample from the hidden representation \(H\) to form \(H'\)\;
		Take a forward step with the GSN using \(H'\) to create \(x'\)\;
		Use this \((x', x)\) pair as a training example for the GSN parameters\;
	}
	\caption{ Walkbacks for sequential input }
\end{algorithm}

\section{RNN-GSN: Generalizing the EM training model for the TGSN}

\begin{figure}[h!]
  \centering
    \includegraphics[width=0.8\textwidth]{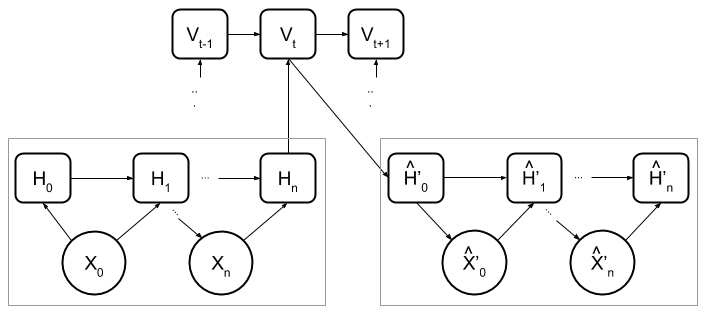}
\caption{Recurrent GSN architecture. H is GSN hiddens, V is RNN hiddens.}
\end{figure}

The EM model is easier to train and appears to have better mixing in both the input and sequence spaces compared to the online learning model. However, due to the simple regression step, it is unable to represent complex sequences in the representation space. A more general model is necessary to encode complex data in both input and representation spaces.

Ultimately, this model generalizes the TGSN by alternating between finding a good representation for inputs and a good representation for sequences. Instead of a direct encoding \(H_{t} \rightarrow H_{t+1}\), this model learns the encoding of \(P(H_{t+1}|H_{t}...H_{0})\) This way, the GSNs can optimize specifically for reconstruction or prediction rather than making the hidden representation learn both. Further, by making the sequence prediction GSN layer recurrent over the top layer of the input reconstruction GSN layer, this system can learn complex, nonlinear sequence representations over the modes of the input space, capturing a very large possibility of sequential data distributions. These two specified layers can then be repeated to form deep, generalized representations of sequential data.

\section{Algorithm}
This algorithm also alternates between training the reconstruction GSN parameters and prediction GSN for transitions.
 \begin{algorithm}[h!]
	\KwIn{training data \(X\) from a sequential distribution \(D\)}
	Initialize reconstruction GSN parameters \(\Theta_{gsn} = \) \{List(weights from one layer to the next), List(bias for layer))\}\;
	Initialize transition RNN parameters \(\Theta_{rnn} = \) \{List(weights from one layer to the next higher), List(weights from one layer to the next lower), List(bias for layer))\}\;
	\While{training not converged}{
		\For{each input \(X\)}{
			Sample from reconstruction GSN with walkback using \(X\) to create \((X_{recon},X)\) pairs for training parameters \(\Theta_{gsn}\)\;
			Compute RNN using the hidden representations \(H\) from the reconstruction GSN on the input \(X\)\;
			Store the predicted next hidden representations \(H'\) and use them with sampling from the next reconstruction GSN to train the transition parameters \(\Theta_{rnn}\)\;
		}
	}
	\caption{ Recurrent GSN Algorithm }
\end{algorithm}

\section{Experimental setup}
The RNN-GSN uses the same general training parameters with regards to noise, learning rate, annealing, momentum, epochs, hidden layers, and activation as the TGSN. In addition, it has one recurrent (LSTM) hidden layer of 3000 units, receiving input from layer 1 and layer 3 of the GSN below it. No sequential walkback steps were performed. The RNN-GSN performed worse with regards to binary cross-entropy of the predicted reconstruction than the TGSN (achieving a score of 0.2695, with the current reconstruction achieving a score of 0.1669). However, the reconstruction and predicted reconstruction after 300 training iterations qualitatively looks like the model is learning the correct sequence. Further, because of the additional recurrent layer and parameters, this model should take longer to train and slower progress to sequence prediction is expected. Further study of this general model should be with hyper-parameter optimization and more training epochs.

\begin{figure}[h!]
  \centering
    \includegraphics[width=1.0\textwidth]{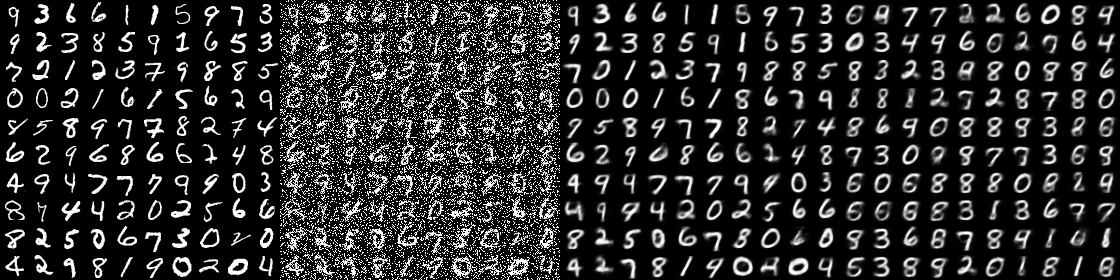}
\caption{RNN-GSN reconstruction of current digits and predicted next digits after 300 iterations.}
\end{figure}

\begin{figure}[h!]
  \centering
    \includegraphics[width=.8\textwidth]{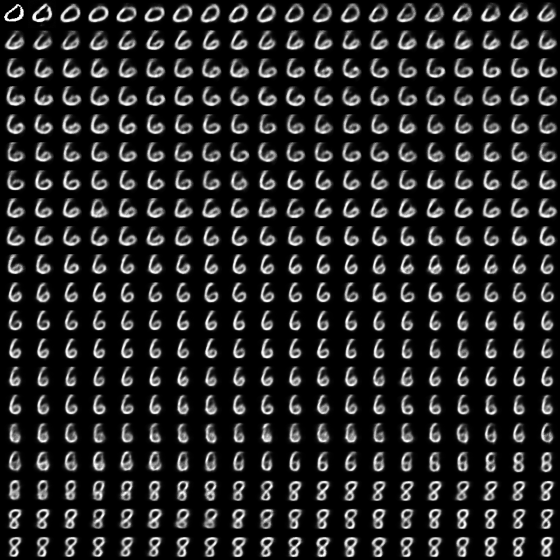}
\caption{RNN-GSN sampling after 300 iterations.}
\end{figure}

%% file: chapters/model3.tex
The TGSN and RNN-GSN models have shown the idea so far of decoupling input representation from sequence representation. However, the sequence complexity learned still has a limit by the RNN representation capacity over the input latent space. We can generalize this decoupling idea even further by creating an alternating structure with these input representation and sequence representation layers, inspired by convolutional neural networks with alternating convolutional and dimensionality reduction layers \cite{lenet5}. The Sequence Encoder Network (SEN) is able to stack these input and sequence representational layers to learn combinations of representations for the sequence dynamics across many layers to enable a much higher capacity for complex inputs.

\section{Algorithm}

\begin{figure}[h!]
  \centering
    \includegraphics[width=0.8\textwidth]{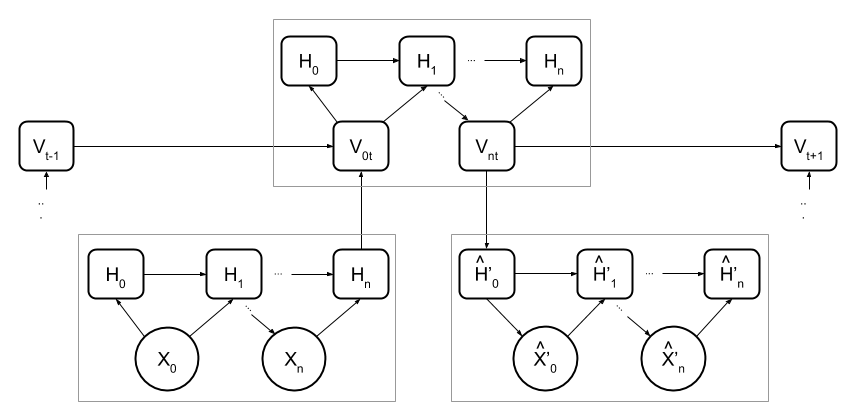}
\caption{Sequence Encoder Network architecture. H is GSN hiddens, V is RNN hiddens.}
\end{figure}

The SEN algorithm extends the RNN-GSN by continuing to learn representations on top of the sequence representations \(V\): 
\begin{enumerate}
\item Use a GSN to learn the generative input representation \(H^0\) of the input \(X\)
\item Use an RNN to learn the sequence representation \(V^0\) over \(H^0\)
\item Use another GSN to learn \(H^1\) over the sequence representations \(V^0\)
\item Use another RNN to \(V^1\) over \(H^1\)
\item Repeat for desired representation layers \(n\) to get top-level sequence representations \(V^n\)
\end{enumerate}

Intuitively, these extra layers enable the network to represent hierarchical sequence dynamics from learning transitions between sequence representation states. This hierarchical property allows for much longer or more complex time series interactions.

The sequence representations can also be interpreted as attractor networks\cite{mathis-mozer} arranged in a hierarchical manner, learning combinations of local sequence states to form global representations. The first GSN over the learned RNN states \(V_0\) forms a localist-attractor module, where the further layered RNN and GSN hidden states reduce the dimensionality and learn increasingly global representations of the sequence state space.

Because we are essentially stacking RNN-GSN layers, the EM approach for training reconstruction and sequences separately would benefit from layerwise pretraining. For the SEN, we combine the forward passes and train both the reconstruction GSN parameters and sequence RNN parameters at the same time to avoid this issue.

While the SEN presented here uses GSN and RNN layers, it can be implemented as any encoder-decoder model that stores hidden state (VAE, convolutional autoencoder, etc.), and then any recurrent model (LSTM, GRU, etc.) to transition between hidden states used by the decoder.

 \begin{algorithm}[h!]
	\KwIn{training data \(X\) from a sequential distribution \(D\)}
	Initialize reconstruction GSN parameters \(\Theta_{gsn_n} = \) \{List(weights from one layer to the next), List(bias for layer))\} for desired layers \(n\)\;
	Initialize transition RNN parameters \(\Theta_{rnn_n} = \) \{List(weights from one layer to the next higher), List(weights from one layer to the next lower), List(bias for layer))\} for desired layers \(n\)\;
	\While{training not converged}{
		\For{each input \(X\)}{
			Run \(X\) through the SEN, creating \(H^0\) to \(H^n\) \(n*(H^i, H`^i)\) reconstruction pairs, and \(H_{t+1}^n\) expected next hiddens from the RNNs \(V^n\)\;
			Calculate the reconstruction loss for \(H\) and prediction loss for \(V\).		}
	}
	\caption{ SEN Algorithm }
\end{algorithm}

%% file: chapters/discussion.tex
The models were evaluated on two standard datasets, videos of bouncing balls and motion capture data, and compared to the RNN-RBM and RTRBM models discussed in Chapter 2 (Related Work) as well as an LSTM.

\section{Samples from RNN-RBM on sequenced MNIST}

Compared to the samples generated by the RNN-RBM, it is clear that the GSN framework has an easier time mixing between modes of the input data. It also appears to form better reconstructions of the input data. This improvement can be attributed to a deeper representation of the input space, since the RNN-RBM only had two layers - one for the RBM and one for the RNN.

\begin{figure}[h!]
  \centering
    \includegraphics[width=0.8\textwidth]{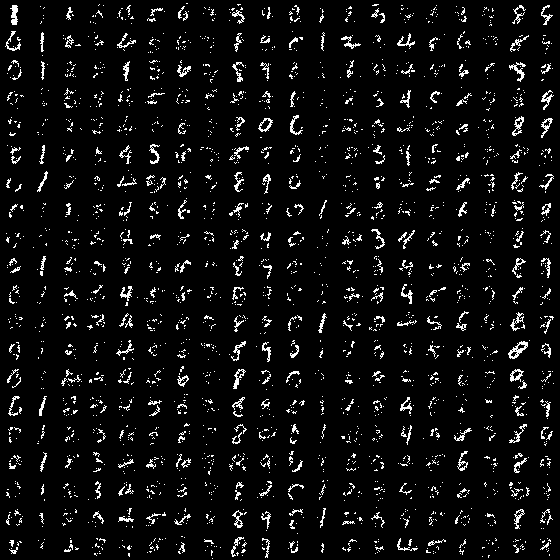}
\caption{RNN-RBM sampling after 300 iterations.}
\end{figure}

\section{Bouncing balls videos dataset}

This dataset generates videos of 3 balls bouncing and colliding in a box as described in \cite{lewandowski12}\footnote{http://www.cs.utoronto.ca/~ilya/code/2008/RTRBM.tar}. The videos have length of 128 frames with 15x15 resolution of pixels in the range of [0, 1]. Training examples are generated artificially at runtime so each sequence seen is unique, which helps reduce overfitting.

The LSTM and Untied GSN models were trained with two layers of 500 hidden units. The Temporal GSN used two layers of 500 units with tied weights, input salt-and-pepper noise of 0.2, hidden Gaussian noise of 0 mean and 1 standard deviation, 4 walkback steps, and a history context window of 4 timesteps. The RNN-GSN had two GSN layers of 500 units with tied weights and 4 walkbacks, and a single layer LSTM with 500 hidden units. The SEN had two GSN layers and two LSTM layers, where the GSN had 2 layers of 500 hidden units with tied weights and 4 walkback steps, and the LSTM had 500 hidden units.

All models were trained on subsequences with length 100 using the Adam optimizer with a learning rate of 0.001, beta1 of 0.9, and beta2 of 0.999. Gradients were scaled to clip the batchwise L2 norm at a maximum of 0.25.

\section{CMU motion capture dataset}

This dataset is a series of captured human joint angles, translations and rotations around the base of a spine as in \cite{sutskever08}\footnote{https://github.com/sidsig/NIPS-2014}. There are 3826 samples of 49 real-valued inputs, so input sampling was not used for the GSN and the visible layer had a linear activation.

The train set was split as the first 80\% of each sequence, with the last 20\% forming the test set.

The LSTM and Untied GSN models were trained with two layers of 128 hidden units. The Temporal GSN used two layers of 128 units with tied weights, input salt-and-pepper noise of 0.1, hidden Gaussian noise of 0 mean and .5 standard deviation, 4 walkback steps, and a history context window of 3 timesteps. The RNN-GSN had two GSN layers of 128 units with tied weights and 4 walkbacks, and a single layer LSTM with 256 hidden units. The SEN had two GSN layers and two LSTM layers, where the GSN had 2 layers of 128 hidden units with tied weights and 4 walkback steps, and the LSTM had 128 hidden units.

All models were trained on subsequences with length 100 using the Adam optimizer with a learning rate of 0.001, beta1 of 0.9, and beta2 of 0.999. Gradients were scaled to clip the batchwise L2 norm at a maximum of 0.25.

\begin{table}[h!]
\begin{tabular*}{\textwidth}{p{4cm} r r r r}
\hlinewd{1.5pt}
  & Bouncing Balls & CMU Motion Capture \\
\hline
LSTM & \bfseries0.11 & 9.24\\
RTRBM & 2.11 & 20.1\\
RNN-RBM & 0.96 & 16.2\\
Untied GSN & 0.94 & \bfseries6.90\\
TGSN & 5.57 & 9.27\\
RNN-GSN & 5.28 & 11.49\\
SEN & 19.0 & 50.8\\
\hlinewd{1.5pt}
\end{tabular*}
\caption{Mean squared prediction error on bouncing balls videos and motion capture data. RTRBM and RNN-RBM numbers from [10]}
\end{table}

\section{Results}
Notably, the baseline LSTM outperformed all other models on the videos of bouncing balls dataset, achieving a mean frame-level square prediction error of 0.11. The Untied GSN had lower error than the RNN-RBM, but the TGSN and RNN-GSN both did much worse. One possible explanation is the EM algorithm entered bad RNN state transitions as discussed in Chapter 5.3. This can be seen in the RNN-GSN frame outputs, which diverged from a good state into a bad representation. Another reason the GSN-based models (except the Untied GSN) performed poorly is the injected salt-and-pepper and gaussian noise remaining relatively high throughout the process. We would like to explore noise scheduling in the future to help training convergence.

\begin{figure}[h!]
  \centering
    \includegraphics[width=0.3\textwidth]{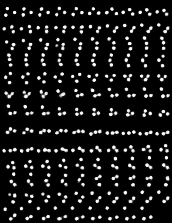}
\caption{RNN-GSN good state.}
\end{figure}

\begin{figure}[h!]
  \centering
    \includegraphics[width=0.3\textwidth]{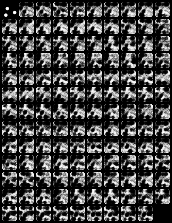}
\caption{RNN-GSN diverged bad state.}
\end{figure}

For the CMU motion capture dataset, the Untied GSN model had the lowest mean frame-level square prediction error at 6.90. The LSTM and TGSN were similar in error and all other models except the SEN outperformed the RTRBM and RNN-RBM baselines. This dataset has a much lower input dimensionality, so we are less likely for the optimization to diverge using the GSN-based models. Further, the added noise for GSNs were lower in this experiment than in the bouncing balls video dataset.

Ultimately the SEN in both experiments was not able to converge. We believe learning reconstructions of higher-level sequence representations without those representations being in a relatively stable starting point leads to high training instability. Future work will explore layer-wise pretraining, or dynamically growing the number of layers to encourage representation stability and training convergence. Further hyperparameter search for learning rate and gradient clipping should also be performed to help stability.

\begin{figure}[h!]
  \centering
    \includegraphics[width=0.3\textwidth]{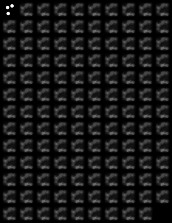}
\caption{SEN frame predictions after 140 epochs.}
\end{figure}

\section{Future Work}
Future work will focus on studying the Sequence Encoder Network class of architectures and their training stability. We would like to explore convolutional autoencoders for image-based prediction tasks, and sequence-to-sequence models for language tasks, with GRU's as recurrent layers. Future work should also explore using adversarial loss during reconstruction to help stability and avoid mode collapse over sequences.

%% file: chapters/conclusion.tex
This thesis presents three models using GSNs to learn useful representations of complex input data sequences. It corroborates that deep architectures, such as the related work with RBMs, are extremely powerful ways to learn complex sequences, and that GSNs are an equally viable framework that improve upon training and inference of RBMs. Deep architectures derive most of their power from being able to disentangle the underlying factors of variation in the input data - flattening the data manifolds at higher representations to improve mixing between the many modes.

The Temporal GSN, an EM approach, takes advantage of the GSN's ability to reduce the complexity of the input data at higher layer of representation, allowing for simple linear regression to learn sequences of representations over time. This model learns to reconstruct both the current input and the next predicted input. This reconstructed predicted input tends to look like an average of the next inputs in the sequence given the current input.

The Recurrent GSN adds a recurrent hidden state to learn a sequential representation between the GSN's latent spaces. This approach allows for more complex time series interactions to be learned over the TGSN.

The Sequence Encoder Network generalizes the idea behind the Recurrent GSN. By alternating layers of encoder-decoder models that learn reconstructions of the input, and recurrent layers that learn reconstruction of future prediction, it models hierarchical representations of both the input and sequence spaces. Training is much more difficult as layer numbers increase.